\documentclass[12pt]{article}
\usepackage[T1]{fontenc}

\usepackage{graphicx}
\usepackage{graphicx} 
\usepackage{amsmath,amssymb}
\usepackage{amsfonts}
\usepackage{tikz}
\usetikzlibrary{shapes.geometric, positioning, backgrounds, calc}
\usepackage{subcaption}
\usepackage{xcolor}
\usepackage{algorithm}
\usepackage{algpseudocode}
\usepackage{listings}
\usepackage{booktabs}
\usepackage{multirow}
\usepackage{makecell}
\usepackage{hyperref}
\usepackage{authblk}

\definecolor{darkgreen}{rgb}{0,0.5,0}  
\lstdefinestyle{customPython}{
  language=Python,
  basicstyle=\ttfamily\scriptsize,
  keywordstyle=\color{blue}\bfseries,
  commentstyle=\color{darkgreen},
  stringstyle=\color{purple},
  numbers=left,
  numberstyle=\tiny\color{gray},
  stepnumber=1,
  breaklines=true,
  frame=single,
  captionpos=b
}

\begin{document}
\title{Exploring Solver-Level Warmstarting
for Neural Network Verification}

\author[1]{Annelot W. Bosman}
\author[2]{Minghao Liu}
\author[2]{Marta Kwiatkowska}
\author[1,3,4]{Holger H. Hoos}
\author[1]{Jan N. van Rijn}

\affil[1]{Leiden Institute of Advanced Computer Science, Leiden}
\affil[2]{University of Oxford, United Kingdom}
\affil[3]{Chair for AI Methodology, RWTH Aachen University, Germany}
\affil[4]{University of British Columbia, Canada}

\date{}
  
\maketitle             
\begin{abstract}
Neural network verification has become a key tool for providing formal guarantees on the behaviour of neural networks.
However, many verification problems remain computationally intractable in the worst case: even for common adversarial robustness specifications, verification is NP-complete. 
Here, we explore the application of solver-level warmstarting for neural network verification to exploit information from previous solutions. 
We study the effect on running time as several properties are modified, 
including perturbation radii, input data and the networks themselves, using a pipeline that is generalisable and potentially adaptable to state-of-the-art verifiers. 
Our results show that warmstarting can significantly reduce verification time in most cases. 
Moreover, warmstarting enables the successful verification of instances that could not be solved from scratch within the given time limit.
\end{abstract}

\section{Introduction}\label{sec:introduction}

Recent advances in computational power and the availability of large-scale data have led to the widespread deployment of neural networks in real-world settings \cite{naude2021artificial}.
As neural networks are adopted in safety-critical domains, such as autonomous driving \cite{zhu2024autonomous} and healthcare \cite{zhou2021reviewmedical}, the demand for formal guarantees in their robustness has become more urgent, since failures can have severe consequences.

Neural network verification aims to determine whether a given neural network satisfies a specified property over a set of inputs, taking the network, input bounds, and property as inputs and returning either a proof that the property holds or a counterexample demonstrating a violation.
Most current neural network verifiers treat each verification task in a stand-alone manner. 
However, in realistic experimental pipelines and in real-world cases, there are many similar verification instances, and it could be beneficial to use these similarities in the verification process.
In the case of adversarial robustness property, it has been common knowledge that networks tend to be misled by similar adversarial examples, which are inputs deliberately modified to exploit model vulnerabilities
\cite{liu2016delving,naseer2019cross,szegedy2013intriguing,waseda2023closer},
and it has been recently shown that similar networks also have similar safe radii for the same input \cite{BosmanEtAl25}.
Incremental constraint solving has been explored for neural network verification to avoid solving each problem from scratch \cite{elsaleh2026incremental,liu2026exact}.
Some work improves cross-instance verification efficiency by processing instances with similar perturbation radii in mini-batches \cite{tzour2025mini} or by exploiting similarities and dependencies between the hidden-layer states of different inputs
\cite{fischer2022shared,banerjee2024input}.
Recent studies also explore relational verification \cite{banerjee2024relational}, which leverages dependencies among multiple executions of the same network,
and incremental verification \cite{ugare2023incremental}, which exploits similarities between the original and updated networks to reuse previous verification effort.
Overall, prior work utilises specific correlations between instances, but is inherently limited in generality;
for example, systematic reuse of information across variations in perturbation radii, input data and networks is not supported.

The verification tasks can be encoded as constraint satisfaction problems and 
solved using mixed-integer linear programming (MILP) solvers.
Warmstarting techniques for MILP have been studied extensively in the mathematical optimisation literature \cite{ralphs2006duality}. 
Given a MILP instance with modified parameters or constraints, 
warmstarting MILP solvers brings gains in efficiency by reusing the information from previous runs on similar instances.

In this paper, we incorporate MILP warmstarting techniques into neural network verification.
We focus on reusing previously obtained solver artefacts, such as previous solutions or the information on search tree, to initialise the solver for previously unseen verification instances.
This set-up enables warmstarting across a wide range of instance variations, largely independently of how the underlying verification property is modified  (e.g., modified perturbation radius), although certain scenarios
are expected to benefit more than others due to better similarity and reusable lower bounds.
Specifically, we combine components from existing verification software with additional interfacing code to construct a pipeline that enables warmstarting at the solver level.
We use SYMPHONY \cite{ralphs2005symphony}, an open-source MILP solver that provides direct support for warmstarting techniques, to establish a controlled verification setting and systematically investigate the impact of solver-level warmstarting.
However, SYMPHONY has not been benchmarked against state-of-the-art MILP solvers for neural network verification, and thus our baseline should not be considered state-of-the-art.
Instead, our goal is to isolate and evaluate the benefits of solver-level warmstarting before integrating these techniques into state-of-the-art verifiers.

The main contributions of this paper are as follows:

\begin{itemize}
    \item To the best of our knowledge, we are the first to study warmstarting for different properties for neural network verification directly at the MILP solver level, in contrast to existing approaches \cite{banerjee2024input,ugare2023incremental} that incorporate warmstarting mechanisms at the verifier level. 
We provide a systematic analysis of how different types of property changes affect the suitability and effectiveness of solver-level warmstarting.
    \item We investigate multiple warmstarting scenarios and strategies, analysing their impact across different property changes.
    Our experimental results demonstrate that this approach can yield significant performance improvements, both in terms of reduced running time and in the ability to solve verification instances that could not be solved without warmstarting within a given time limit.
     Importantly, these improvements are observed not only for small property changes, such as changes in the perturbation radius~$\varepsilon$, but also for different inputs and neural networks.
    \item We present a proof-of-concept pipeline and empirical evaluation of solver-level warmstarting using the SYMPHONY solver \cite{ralphs2005symphony}, demonstrating its potential benefits and limitations in practice.
\end{itemize}

Our experimental code and results can be found here:

\small{\url{https://github.com/ADA-research/STAI-Solver-Level-Warmstarting-paper}}.

\section{Problem Specification}\label{sec:problem}

In this section, we first introduce the formalisation of neural networks and the verification tasks at hand.
Next, we outline the scenarios we considered in which warmstarting would be beneficial in verification.

\subsection{Verification of Neural Networks}
A deep neural network classifier is a function
    $f_{\theta} : \mathbb{R}^{n_0} \rightarrow \mathbb{R}^{n_{L}}$,
where $\theta$ denotes the trainable parameters, $n_0$ is the input dimension, and $n_{L}$ is the number of output classes.
The network consists of $L$ layers, with layer $l$ containing $n_l$ neurons.

In the following, we consider fully connected neural networks (FCNNs), where the structure of layer $l$ can be formalised as:
\begin{equation}
    x_{l}=\sigma\left( W_l \cdot x_{l-1} + b_l \right),
\end{equation}
where $W_l \in \mathbb{R}^{n_l \times n_{l-1}}$ and $b_l \in \mathbb{R}^{n_l}$ are parameters, and $\sigma$ is an activation function, 
which in our study we assume to be the widely used ReLU function $\sigma(x)=\max (x,0)$.

Formal verification utilises mathematically rigorous methods to determine whether a neural network satisfies a specific property for all inputs within a given input region.
Assume a feasible region $G$ and a property $P$ in the form of a Boolean formula over linear inequalities.
Following \cite{de2021improved}, formal verification aims to decide whether

\begin{equation}
      \forall x_0 \in G : \quad \left( x_{L} = f_\theta(x_0) \right) \;\Longrightarrow\; P(x_{L}) \label{eq:property},
\end{equation}
where $x_L$ is the output vector of $f_\theta$.
In this work, we are concerned with the \emph{adversarial robustness} property, which is a key property for neural networks and has been extensively studied in the literature \cite{meng2022adversarial}.

Given an original input $x^*$ with correct label $\lambda(x^*)$, we consider the feasible region is defined by an $L_\infty$-norm ball centred at $x^*$ with a radius of $\varepsilon$:
\begin{equation}
    G_{\varepsilon}(x^*) = \left\{ x_0: |x^* - x_0|_\infty \leq \varepsilon \right\},
\end{equation}
where ${|x|}_\infty = \max_{1 \le i \le n}{|x_i|}$.
A neural network $f_\theta$ is said to be robust at $x^*$ if, and only if, 
$\forall x_0 \in G_{\varepsilon}(x^*):  \arg \max_if_\theta(x_0)[i]= \lambda(x^*)$.

The robustness verification of FCNNs can be formulated as the following mixed-integer linear programming (MILP) problem:

\begin{subequations} \label{eq:milp-encoding}
\begin{align}
   \max \quad & y & \label{eq:milp-encoding-obj} \\
    \text{s.t.} \quad & y = \max_{1 \le i \le n_{L}, i \neq \lambda(x^*)}{x_{L}[i] - x_{L}[\lambda(x^*)]} \label{eq:feasible0} & \\
    & \hat{x}_l = W_l \cdot x_{l-1} + b_l & \forall l \in [1,L] \label{eq:feasible1} \\
    & x_l = \max \left( \hat{x}_{l},\,0 \right) & \forall l \in [1,L] \label{eq:feasible2} \\
    & x^*-\varepsilon \le x_0 \le x^* + \varepsilon, & \label{eq:feasible3}
\end{align}
\end{subequations} 
where $x_1, \dots, x_L$ are vectors of real-valued variables.
Note that the $\max$ functions can be represented exactly with linear constraints using the big-M method \cite{tjeng_evaluating_2019}.
A neural network is robust at $x^*$ with perturbation radius $\varepsilon$ if, and only if, the optimal value of Objective~(\ref{eq:milp-encoding-obj}) is less than 0, which corresponds to the satisfaction of the property in Eq.~(\ref{eq:property}).

\subsection{Scenarios for Warmstarting}\label{sec:scenarios}

In this paper, we always consider two properties: the original property $P_x$, with an associated
solution $A_x$ (whose concrete form we discuss later), 
and a new property $P_z$.
There are multiple ways in which $P_x$ and $P_z$ may be related to one another, and understanding
these relationships is key to determining how $A_x$ can be reused when verifying $P_z$.

\subsubsection{Different \texorpdfstring{$\varepsilon$}{epsilon}.}

A first way in which $P_z$ may differ from $P_x$ is through a change in the perturbation radius. 
This change is relevant when the feasible region of the verification problem depends on a predefined $L_\infty$-norm ball with a radius of $\varepsilon$. 
Given the same reference input $x^*$, we now consider two feasible region sets.
\[
    G_{\varepsilon_x}(x^*),G_{\varepsilon_z}(x^*) \quad \text{with} \quad \varepsilon_z \neq \varepsilon_x.
\]
This change only has an effect on Constraint (\ref{eq:feasible3}), as this encodes the feasible region of the modified input.
While this is the simplest type of change we consider, it is particularly relevant for finding the largest safe perturbation radius for an input to a given network. Running time tends to become prohibitively expensive near the decision boundary \cite{BosmanEtAl25}, so reducing running time on previously unseen instances can help obtain a clearer picture of the robustness of a given network. 

We consider two warmstarting scenarios.
If $\varepsilon_z < \varepsilon_x$, then $G_{\varepsilon_z}(x^*) \subseteq G_{\varepsilon_x}(x^*)$,
and the feasible region for $P_z$ is a strict subset of that for $P_x$.
In such cases, all artefacts in solution $A_x$ remain valid, although they could be an over-approximation. 
Conversely, if $\varepsilon_z > \varepsilon_x$, the new perturbation region expands, potentially invalidating some parts of $A_x$.
Cuts and bounds derived under $P_x$ may no longer hold, and branch-and-bound nodes previously ruled out might need to be reopened.
This scenario, therefore, provides a natural test of robustness for warmstarting strategies.
We note that, in the trivial case where $\varepsilon_z > \varepsilon_x$ and $P_x$ was violated, $P_ z$ will also always be violated.

\subsubsection{Different \texorpdfstring{$x^*$}{input}.}

Another change in property arises when the reference input is replaced by a new input $z^* \neq x^*$.
The verification property becomes
\[
    P_z: \quad \exists\, z_0 \in G_{\varepsilon}(z^*) \quad \text{s.t.}\; \arg\max_i f_\theta(z_0)[i] \neq \lambda(z^*).
\]
This change in property affects Constraints (\ref{eq:feasible1}) and (\ref{eq:feasible3});
as a result, some artefacts in $A_x$ (e.g., neuron phase assignments, linear relaxations,
interval bounds) may no longer be valid without modification.

Nonetheless, $P_x$ and $P_z$ can still be closely related when $z^*$ lies near $x^*$, for example, when they are from the same class. 
In this case, much of the internal activation pattern may be preserved,
and several components of $A_x$ can be meaningfully warmstarted for $P_z$ after appropriate validity checks.

\subsubsection{Different \texorpdfstring{$f_{\theta}$}{model}.}

A more substantial change occurs when network parameters or architectures are modified.
If the original model is $f_{\theta}$ and the new model $f'_{\theta'}$ with $\theta' \neq \theta$, and $f$ is allowed to be different from $f'$, the property becomes
\[
    P_z: \quad \exists\, x_0 \in G_{\varepsilon}(x^*) \quad \text{s.t.}\; \arg\max_i f'_{\theta'}(x_0)[i] \neq \lambda(x^*).
\]
In this case, Constraint (\ref{eq:feasible1}) may be affected, which concerns the vast majority of the encoding; even small perturbations to weights or biases can change activation patterns and thus
alter the feasible region of the verification problem.

If the change in parameters is small (e.g., fine-tuning, pruning or lightweight retraining),
many artifacts in $A_x$ may remain approximately valid. 
In such cases, warmstarting remains possible but requires conservative checks or
adjustments (e.g., inflating abstract bounds, re-validating cuts).
However, if the network structure changes significantly (e.g., new layers, different activations),
then $A_x$ may no longer be reusable.
 
\section{Methodology}\label{sec:method}

In this section, we first introduce the core theory and algorithms of warmstarting in MILP solving.
Next, we describe our approach of applying warmstarting techniques in MILP-based neural network verification.

\subsection{Warmstarting in MILP Solving}

Consider a MILP instance
\[
    z(b):=\min_{x \in \Pi(b)}{c^\top x}, \qquad \Pi(b) = \left\{ x \in \mathbb{Z}^p \times \mathbb{R}^{n-p} \mid A \cdot x=b, x \ge 0 \right\},
\]
where $x$ is a vector of $n$ variables,
with $p$ of them being integers, and $c,A,b$ are parameters.

To determine the optimal value of $z(b)$, a common procedure is the branch-and-bound algorithm, which constructs a search tree with node set $V$ by recursively partitioning the feasible region into subproblems.
Each node $i \in V$ of the tree corresponds to a restricted feasible region $\Pi_i(b) \subseteq \Pi(b)$ by imposing additional bounds on integer variables.
At each node, a lower bound $l_i(b)$ is obtained by solving the following linear programming (LP) relaxation $l_i(b) := \min_{x \in P_i(b)}{c^\top x}$ with the simplex algorithm \cite{dantzig1951maximization}, where $P_i(b)$ is a polyhedral relaxation of $\Pi_i(b)$.
An optimal basis $B_i$, consisting of a set of linearly independent columns in $A$, together with the assignments of the non-basic variables fixed at their bounds, is determined.
The global lower bound can be maintained by $L=\min_{i \in V}{l_i(b)}$.

Consider a MILP with modified right-hand side $d$ and objective $z(d)$.
The idea of warmstarting is to solve this instance by using artefacts (e.g., the search tree, lower bounds, and cutting planes) obtained during the solution of $z(b)$.
In this paper, we introduce the warmstarting techniques proposed by \cite{ralphs2006duality}, which provide a dual interpretation of branch-and-bound.
At a given node $i$, suppose the LP relaxation admits an optimal basis $B_i$;
then the optimal LP value can be expressed as an affine function of the right-hand side:
\[
l_i(d) = c_{B_i}^\top \cdot B_i^{-1} \cdot d + \beta_i,
\]
where $c_{B_i}$ denotes the components of $c$ corresponding to the columns of $B_i$, and $\beta_i$ represents the constant factors associated with the non-basic variables. This is a valid dual bound for the MILP, since for any $d$, $l_i(d) \le z(d)$.
Therefore, the piecewise linear function $F(d) = \min_{i \in V}{l_i(d)}$ represents a global lower bound of $z(d)$ and can be used to warmstart the branch-and-bound tree for any new $d$.
Moreover, the pool of global cutting planes is also kept and can be reused for different values of $d$.

We note that, when $A$ is modified, the artefacts that can be reused to solve the new MILP problem are limited.
The search tree structure can be kept, but the dual bounds and global cuts may no longer be valid, and the LP relaxation needs to be recomputed.
Thus, the effectiveness of warmstarting is reduced compared to modifying the parameters on the right-hand side.

\subsection{Neural Network Verification with Warmstarting}\label{sec:warmstarting_pipeline}

In this work, we use the {SYMPHONY} MILP solver \cite{ralphs2005symphony} to perform warmstarting, since, as far as we are aware, it is the only open-source MILP solver with native warmstarting support\footnote{While proprietary solvers (e.g., CPLEX and Gurobi) support warmstarting, the reused artefacts are not publicly documented, which is important in our setting.}.
While SYMPHONY has primarily been applied in the context of operations research, it has not, to the best of our knowledge, yet been explored for neural network verification. 
Importantly, SYMPHONY provides native support for two warmstarting methods that are particularly relevant in our setting: (i) reusing the branch-and-bound search tree (warmstart tree), and (ii) reusing the pool of cutting planes generated for a previous instance. 

\begin{figure}[t]
    \centering
    \includegraphics[width=0.9\linewidth]{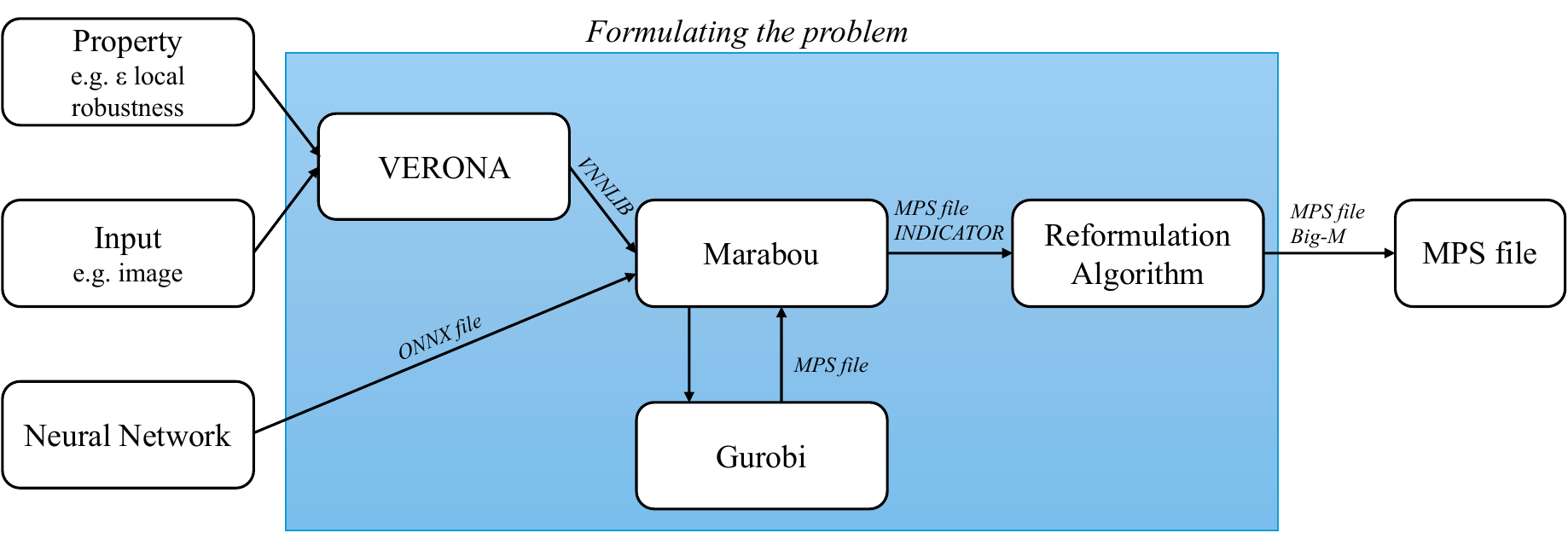}

    \caption{\small The pipeline created to formulate the verification tasks as MPS files that can be used as inputs for the SYMPHONY solver.}
    \label{fig:pipeline_mps}

\end{figure}

In order to use SYMPHONY for neural network verification, we need to generate an MPS file containing the MILP formulation of the network and the property to be verified. 
The complete pipeline of our MPS generation process is depicted in Figure~\ref{fig:pipeline_mps}.

To generate this formulation, we build on the Marabou verifier \cite{katz2017reluplex}, which can encode neural network verification problems as MILPs by interfacing with the Gurobi MILP solver.
Marabou takes as input a neural network and a corresponding VNN-LIB specification file that encodes the input constraints and the property to be verified.
In our workflow, these VNN-LIB files are generated using VERONA \cite{BosmanEtAl25}, which we use as our experiment manager (see Section~\ref{sec:experimental_setup}).
We modified Marabou to terminate the process after Gurobi exports the corresponding MPS formulation, rather than solving the instance directly.
Connecting all these tools to work together smoothly required substantial changes. 
On the VERONA side, we created an interface for SYMPHONY and Marabou; 
on the Marabou side, we made sure that MPS files could be retrieved without interfering with the C compiler. 
For additional details on how we configured and changed these tools, we refer the interested reader to Appendix~\ref{app:tools}.

Modern neural network verifiers represent the result of substantial engineering effort and incorporate a wide range of advanced techniques, including sophisticated branching heuristics, tight relaxations, and parallel solving strategies.
While we believe that such state-of-the-art verifiers could also benefit from warmstarting, directly integrating and evaluating warmstarting strategies within these systems would require significant additional engineering effort and may obscure the fundamental effects of warmstarting itself.
Instead, we adopt SYMPHONY as a controlled proof-of-concept solver that allows us to systematically study warmstarting behaviour in isolation.
This separation of warmstarting from verifier-specific optimisations enables a more transparent
analysis of when and why warmstarting is beneficial.

\subsection{Reformulation Algorithm}

A practical challenge is that the exported MPS formulation contains indicator constraints, which are not supported by SYMPHONY. To address this issue, we introduce a reformulation procedure %
that converts indicator constraints into an equivalent MILP formulation, enabling SYMPHONY to read and solve the resulting instance.

The original purpose of the indicator constraints in our MPS formulation is to encode the \emph{classification decision} of the network, i.e., to enforce that exactly one output class is selected. Listing~\ref{code:indicators} provides a simplified example illustrating how these indicator constraints appear in the exported MPS file.

The mathematical interpretation of the indicator constraint in Listing~\ref{code:indicators} is shown in Eq.~(\ref{eq:indicators}). 
In this formulation, the output variable for a particular class, $x_1$, is controlled by a binary variable $a_1$. 
In the full verification problems, multiple classes are present, and an additional constraint ensures that exactly one class is predicted, typically by enforcing a one-hot structure over the binary variables (e.g., $\sum_{i \in n_L} a_i = n_L - 1$). 
We omit these additional constraints here for simplicity.

\begin{lstlisting}[float=t,style=customPython, caption={Example MPS formulation containing an indicator constraint.}, label={code:indicators}]
ROWS: 
    L GC0 
COLUMNS:
    x1   GC0 1
RHS:
    RHS GC0 0
BOUNDS:
  BV BND1  a1
  UP BND1  x1  m0
  LO BND1  x1  0
INDICATORS:
    IF GC0 a1 1 
ENDATA
\end{lstlisting}

\begin{equation}
x_1 =
\begin{cases}
0, & \text{if } a_1 = 1, \\
\geq 0, & \text{if } a_1 = 0.
\end{cases}\label{eq:indicators}
\end{equation}

\begin{lstlisting}[style=customPython, caption={Example MPS formulation containing a big-M formulation}, label={code:bigM}]
ROWS: 
    L GC0 
    G N1
COLUMNS:
    x1   GC0 1
    x1   N1  1
    a1   GC0 m0
RHS:
    RHS GC0 m0
    RHS N1  0
BOUNDS:
  BV BND1  a1
  UP BND1  x1  m0
  LO BND1  x1  0
ENDATA
\end{lstlisting}

We reformulate the indicator constraint using big-M constraints, as shown in Eq.~(\ref{eq:bigM}) and~(\ref{eq:pos}).
Moreover, we select $M$ as the upper bound of the corresponding variable, i.e., $M = m_i$. 
This choice yields the smallest valid value of $M$ while ensuring that Eq.~(\ref{eq:bigM}) and~(\ref{eq:pos})
remain correct reformulations of Eq.~(\ref{eq:indicators}). 
The resulting MPS file after reformulation is shown in Listing~\ref{code:bigM}.
\begin{align}
    x_1 +M \cdot a_1 &\leq M \label{eq:bigM} \\
    x_1 &\geq 0 \label{eq:pos}
\end{align}
Note that the performance of the MILP solver can depend on the tightness of the big-M encoding.
Tighter variable bounds can strengthen the linear relaxation and reduce search effort.
Since state-of-the-art verifiers typically employ more advanced bound-tightening techniques than those considered here, their MILP formulations may exhibit different solving behaviour.
The impact of such tighter formulations on the effectiveness of warmstarting remains an interesting direction for future work.

The entire reformulation procedure is shown in Algorithm \ref{alg:mps_reformulation}. 
\begin{algorithm}[H]
\caption{MPS Reformulation: \texttt{INDICATORS} $\rightarrow$ big-$M$ constraints}
\label{alg:mps_reformulation}
\begin{algorithmic}[1]
\Require Input MPS file $F_{\text{in}}$ containing an \texttt{INDICATORS} section
\Ensure Output MPS file $F_{\text{out}}$ containing no indicator constraints

\State Parse $F_{\text{in}}$ into list of lines $S$
\State Identify section boundaries in $S$ (e.g., \texttt{ROWS}, \texttt{COLUMNS}, \texttt{RHS}, \texttt{BOUNDS}, \texttt{INDICATORS}, \texttt{ENDATA})
\State $c \gets 0$

\ForAll{indicator constraints of the form \texttt{IF} $(r, a)$ in \texttt{INDICATORS}}\\
    \Comment{$r$: affected row, $a$: binary indicator variable}
    
    \State Determine the affected continuous variable $x$ associated with row $r$\\ by scanning \texttt{COLUMNS}
    \State Extract an upper bound $M$ for $x$ from the \texttt{BOUNDS} section
    \If{no valid upper bound exists}
        \State Set $M \gets 100$ \Comment{fallback used in our implementation}
    \EndIf
    
    \State \textbf{RHS adjustment:} update RHS entry for $r$ as $b_r \gets b_r + M$
    \State \textbf{Column insertion:} add a coefficient $M$ for variable $a$ in row $r$
    \State \textbf{Row type update:} set the type of $r$ to \texttt{L} (i.e., $\leq$)
    
    \State \textbf{Non-negativity constraint:}
    \State Create a new row $N_c$ of type \texttt{G} (i.e., $\geq$)
    \State Add RHS entry $b_{N_c} \gets 0$
    \State Insert coefficient $1$ for variable $x$ in row $N_c$
    
    \State $c \gets c + 1$
\EndFor

\State Remove the entire \texttt{INDICATORS} section from $S$
\State Write the modified lines $S$ to $F_{\text{out}}$
\end{algorithmic}
\end{algorithm}

\section{Setup of Experiments}\label{sec:experimental_setup}

In this section, we describe the experimental setups, including
details on how we selected the source and target instances for warmstarting in our empirical study.

\subsection{Network and Image Selection}

We selected three fully connected neural networks, mnist-net, mnist-net\_256x2, and mnist-net\_256x4, commonly used in the neural network verification literature. These networks are trained on the MNIST dataset \cite{LeCunEtAl1998} and are relatively small, allowing us to keep computational costs manageable while studying the effect of warmstarting.

\begin{table}[htbp]
\centering
\setlength{\tabcolsep}{4pt}
\begin{tabular}{lrrrr}
\toprule
Network            & SAT & UNSAT & ERROR & TIMEOUT \\ \midrule
mnist-net          & 10  & 104   & 2     & 144     \\
mnist-net\_256x2   & 26  & 118   & 1     & 138     \\
mnist-net\_256x4   & 0   & 63    & 0     & 192     \\
\bottomrule
\end{tabular}

\vspace{0.5em}
\caption{Details on the number of verification instances per network, grouped by solver outcome (SAT, UNSAT, ERROR, and TIMEOUT).}
\label{tab:baseline_results}
\end{table}

We randomly selected 10 test images from the MNIST dataset.
Note that not all selected images were used for every network, since we only considered instances that were classified correctly by the corresponding network.

\subsection{Baseline Instance Generation}
To generate baseline results for each image--network combination to which we can compare our method, we first performed a coarse search over the perturbation radius $\varepsilon$. Specifically, we did grid search \cite{BosmanEtAl25} for $\varepsilon \in [0.001, 0.4]$ with a step size of $0.02$, resulting in 20 candidate values. 
Next, we refined the search around the transition region between UNSAT and SAT by performing two iterative searches with a step size $0.002$.

We refine the boundary from both sides: starting from the largest value for which the property still held, we increased $\varepsilon$, and starting from the smallest value for which the property did not hold, we decreased $\varepsilon$, in both cases using a step size of 0.002 until a timeout is reached.
This staged procedure allowed us to obtain a denser set of baseline measurements near the decision boundary while keeping the total computational cost manageable, because each individual verification query was subject to a 1 wall-clock hour timeout, and we observed a substantial number of timeouts -- particularly in the region where instances transition from UNSAT to SAT. 
Table~\ref{tab:baseline_results} shows the number of generated baseline instances for each solver outcome.

\subsection{Warmstarting Instance Selection}

In this work, we investigate six different warm-starting scenarios. Each scenario consists of a \emph{source} instance and a \emph{target} instance. 
The source instance is solved first and results in either a SAT or UNSAT outcome, while the target instance is, in principle, unsolved at the moment warmstarting is applied. 
In practice, however, we solve all instances independently as well, in order to obtain a baseline running time for each verification task, in order to be able to quantify the potential benefit of warmstarting.
Each of these warmstarting scenarios is discussed in Section \ref{sec:scenarios}.
We now describe how we select the different instances considered for warmstarting.

For each network--image combination, we consider the set of perturbation radii $\varepsilon$ for which the baseline verification outcome is UNSAT and the set for which the outcome is SAT. 
If there are at least two UNSAT instances, we generate a warmstart experiment for \emph{every unique pair} of UNSAT radii $(\varepsilon_a, \varepsilon_b)$, such that $\varepsilon_a < \varepsilon_b$. 
Each pair defines a source and target instance taken from the corresponding MILP encodings.
We repeat the same procedure for the SAT set: if at least two SAT instances exist, we create a warmstart experiment for \emph{every unique pair} $(\varepsilon_a, \varepsilon_b)$ such that $\varepsilon_a > \varepsilon_b$.

After this, again for each network--image combination, we construct a sorted list of all evaluated perturbation radii $\varepsilon$. 
For every $\varepsilon$ value that corresponds to a timeout, we select a \emph{source} instance by finding the nearest $\varepsilon$ for which a solved baseline result (SAT or UNSAT) is available. 
We then pair this solved instance with the timeout instance at the original $\varepsilon$, creating a warmstart experiment where the solved instance acts as the source and the timeout instance acts as the target.

For a fixed network $f$ and target image $x$, we iterate over the available $\varepsilon$ values for $(f,x)$.
For each $\varepsilon$, we select a source instance from the \emph{same network} $f$ but a \emph{different image} $x' \neq x$, provided that an instance for $(f,x',\varepsilon)$ exists.
We then create a warm-started experiment in which the target instance is $(f,x,\varepsilon)$ and the source instance is $(f,x',\varepsilon)$, and we label this experiment as IMAGE.

To construct network warmstart experiments, we instead fix an image $x$ and an $\varepsilon$ value, and collect all instances across networks that correspond to $(x,\varepsilon)$.
If at least two different networks provide an instance for the same $(x,\varepsilon)$ pair, we form an experiment by pairing the first two such instances (with identical $(x,\varepsilon)$ but different networks). We label this experiment as NET.

\subsection{Execution Environment}
All experiments were carried out on a cluster of machines equipped with Intel Xeon Gold 6252 @ 2.10GHz CPUs with 96 cores, 16GB cache size and 252GB RAM, running Ubuntu 24.04 OS.

\begin{table}[t]
\centering
\setlength{\tabcolsep}{4pt}
\renewcommand{\arraystretch}{1.1}
\begin{tabular}{llrrrrr}
\toprule
Network & Category & \#Inst & Better & Worse & Extra & ERR \\
\midrule

\multirow{3}{*}{mnist-net}
   & IMAGE               & 79  & 36  & 12  & 12  & 0  \\
                                  & $\varepsilon$:U--TO & 30  & 30  & 0   & 30  & 0  \\
                                  & $\varepsilon$:U--U  & 968 & 786 & 9   & 0   & 24 \\
\midrule

\multirow{4}{*}{mnist-net\_256x2}
& IMAGE               & 113 & 88  & 16  & 28  & 0  \\
                                  & $\varepsilon$:S--S  & 120 & 4   & 116 & 0   & 0  \\
                                  & $\varepsilon$:S--TO & 21  & 21  & 0   & 21  & 0  \\
                                  & $\varepsilon$:U--U  & 648 & 564 & 12  & 0   & 35 \\ 
\midrule

\multirow{3}{*}{mnist-net\_256x4}
 & IMAGE               & 39  & 26  & 2   & 12  & 0  \\
                                  & $\varepsilon$:U--TO & 140 & 140 & 0   & 140 & 0  \\
                                  & $\varepsilon$:U--U  & 164 & 94  & 2   & 0   & 8  \\ \midrule
                                  & NET                 & 133 & 54  & 43  & 12  & 2 \\
\bottomrule
\end{tabular}

\vspace{0.5em}
\caption{Experimental results obtained from using SYMPHONY for warmstarting neural network verification. 
The results are split per network and category, where we summarised the category names using S(SAT), U (UNSAT), and TO(Timeout). 
We report the number of instances that could not be solved before using warmstarting (Extra); we also report the number of instances for each category that led to an error when using warmstarting (ERR), the total number of instances without considering instances that led to an error (\#Inst) and the number of instances that were solved in less running time including extra instances (Better).}
\label{tab:results-counts}
\end{table}

\section{Results}\label{sec:results}

In this section, we present the results of using SYMPHONY to warmstart new MILP formulations that are slight to substantial modifications of previously solved verification instances. 
The absolute number of evaluated instances, as well as statistics such as the number of warmstart errors and the number of instances with reduced running time, are reported in Table~\ref{tab:results-counts}.

Moreover, Table~\ref{tab:results-times} reports the running times measured in our experiments.

\begin{table}[t]
\centering

\setlength{\tabcolsep}{5pt}
\renewcommand{\arraystretch}{1.1}
\begin{tabular}{llrllr}
\toprule
Network & Category & \# inst & Baseline (s) & Warm (s)  & $\Delta t$ (\%) \\
\midrule

\multirow{3}{*}{mnist-net}
  & IMAGE                & 79  & 1333 ± 1491       & \textbf{218 ± 438} & -83.60 \\
                                  & $\varepsilon$: U--TO & 30  & 3600 ± 0.00           & \textbf{877 ± 255} & -75.60\\
                                  & $\varepsilon$: U--U  & 968 & 538 ± 859          & \textbf{16 ± 75}   & -97.00 \\
\midrule

\multirow{4}{*}{mnist-net\_256x2}
& IMAGE                & 113 & 1635± 1619       & \textbf{295 ± 435} & -82.00 \\
                                  & $\varepsilon$: S--S  & 120 & \textbf{142 ± 206} & 731 ± 365        & 415.50 \\
                                  & $\varepsilon$: S--TO & 21  & 3600 ± 0.00          & \textbf{512 ± 388} & -85.80 \\
                                  & $\varepsilon$: U--U  & 648 & 730 ± 912          & \textbf{83± 227}  & -88.60 \\
\midrule

\multirow{3}{*}{mnist-net\_256x4}
 & IMAGE                & 39  & 1804 ± 1632       & \textbf{65 ± 142}  & -96.40 \\
                                  & $\varepsilon$: U--TO    & 140 & 3600 ± 0.00           & \textbf{341 ± 377} & -90.50 \\
                                  & $\varepsilon$: U--U     & 164 & 503 ± 664          & \textbf{21 ± 88}   & -95.80 \\
 \midrule
& NET                  & 133 & 978 ± 1299                 & \textbf{350 ± 502}         & -64.20     \\
\bottomrule
\end{tabular}

\vspace{0.5em}
\caption{\small Empirical results obtained from using SYMPHONY for warmstarting neural network verification.
The results are shown per network and type of warmstarting (e.g., warmstarting across images, networks and values for $\varepsilon$), where we further split the experiments involving warmstarting across $\varepsilon$-values into variations of S(SAT), U (UNSAT), TO(Timeout).
We report the average baseline and warmstarted running time in seconds; 
we also report the relative change in running time as percentages. 
}

\label{tab:results-times}
\end{table}

\subsection{Different \texorpdfstring{$\varepsilon$}{epsilon}.}
\begin{figure}[!t]
    \centering
    \begin{subfigure}{0.48\linewidth}
        \centering
        \includegraphics[width=\linewidth]{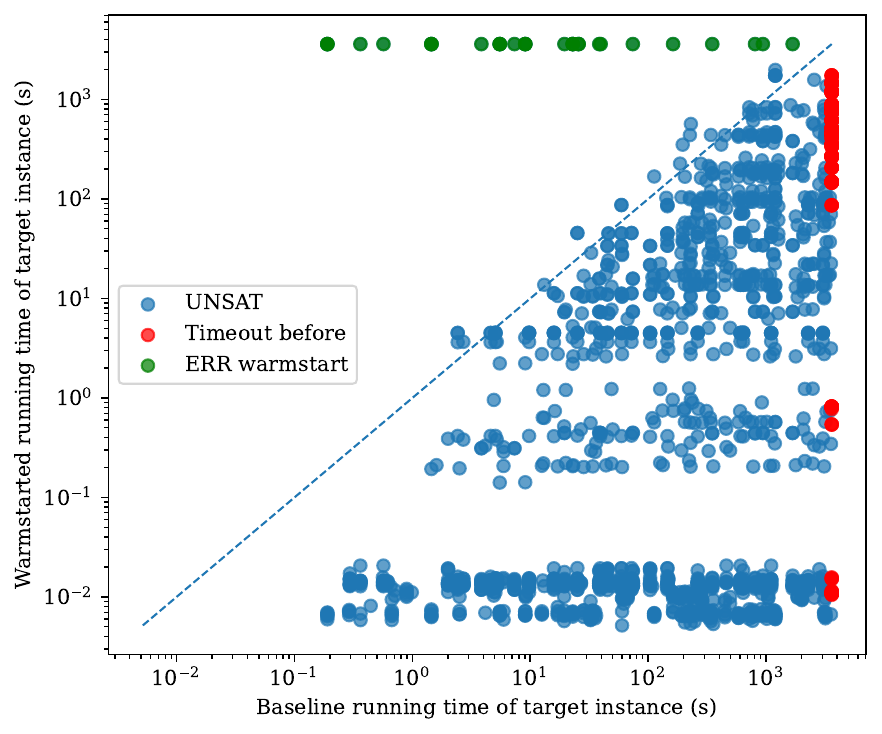}
        \caption{ \small Warmstart for different $\varepsilon$ on UNSAT source and UNSAT or TIMEOUT target instances.}
        \label{fig:unsat_scatter}
    \end{subfigure}
    \hfill
    \begin{subfigure}{0.48\linewidth}
        \centering
        \includegraphics[width=\linewidth]{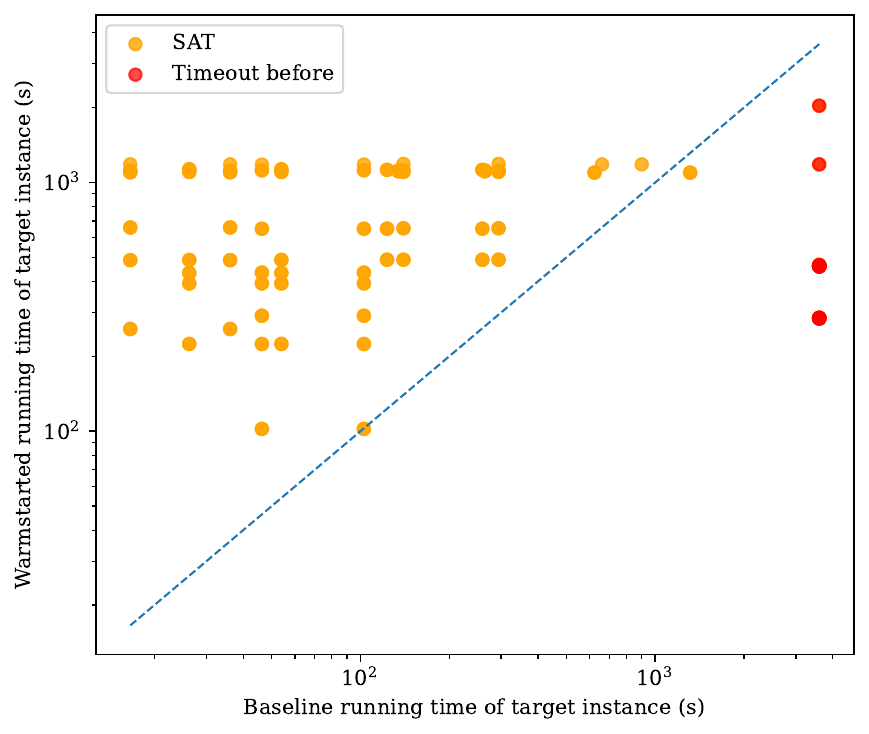}
        \caption{\small Warmstart for different $\varepsilon$ on SAT source and SAT or TIMEOUT target instances.}
        \label{fig:sat_scatter}
    \end{subfigure}
    \begin{subfigure}{0.48\linewidth}
        \centering
        \includegraphics[width=\linewidth]{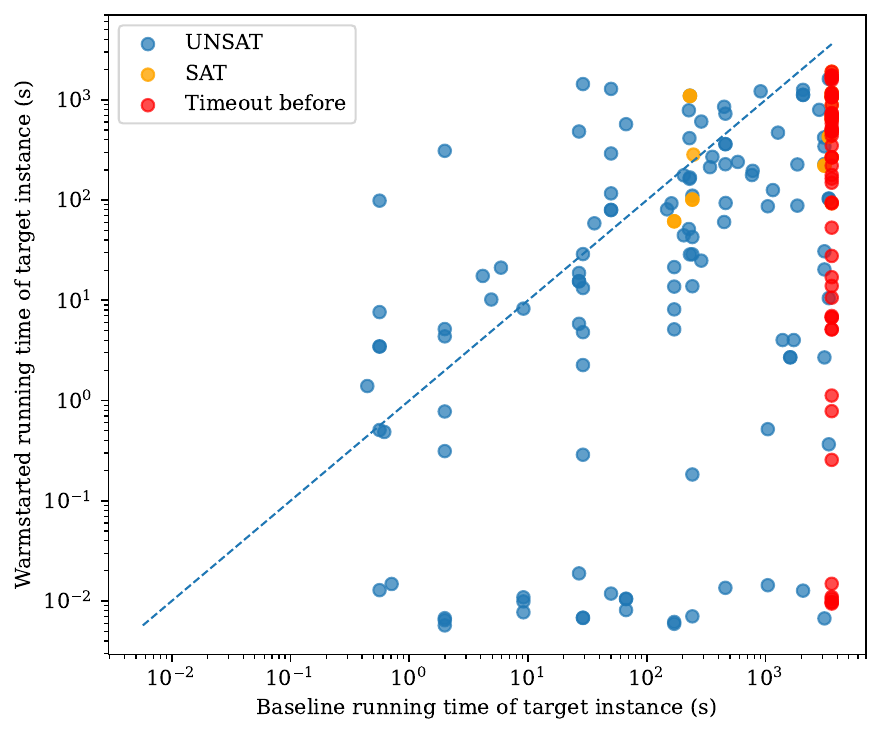}
        \caption{\small Warmstart for the same $\varepsilon$ and network, but different images as source and target instances.}
        \label{fig:image_scatter}
    \end{subfigure}
    \hfill
    \begin{subfigure}{0.48\linewidth}
        \centering
        \includegraphics[width=\linewidth]{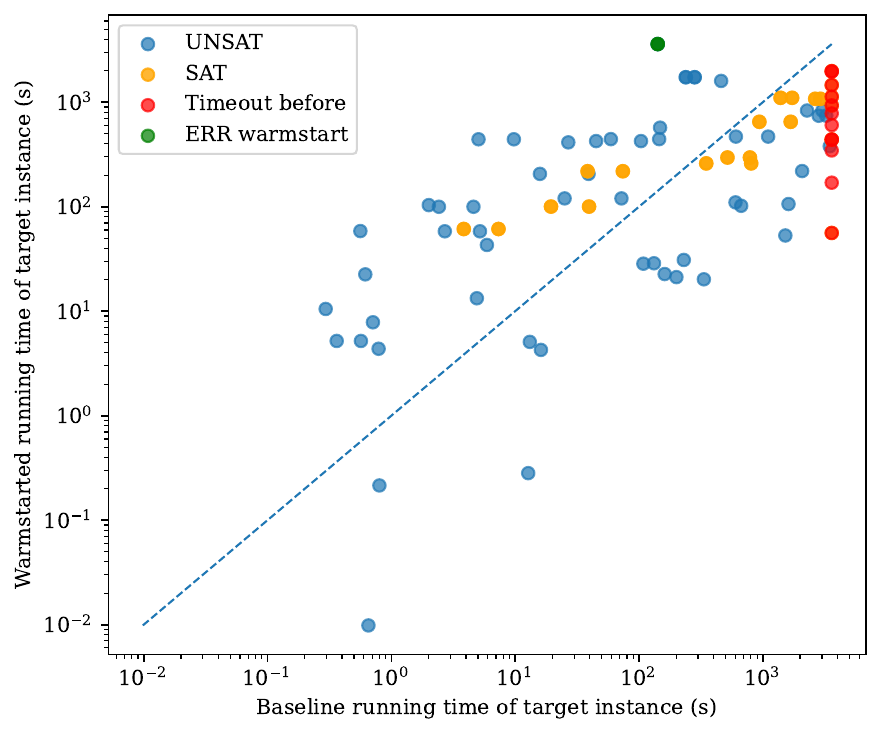}
        \caption{\small Warmstart for the same $\varepsilon$ and image, but different networks as source and target instances.}
        \label{fig:network_scatter}
    \end{subfigure}
    \caption{\small Scatter plots comparing the running times of baseline MILP instances and the corresponding warmstarted instances.
    Warmstarted instances that were solved normally with UNSAT and SAT source are coloured blue in the top two figures, and in the bottom two figures, we differentiate the SAT and UNSAT instances by colouring the SAT instances orange. 
    Instances that were previously leading to a timeout error but now could be solved are coloured red, and instances that led to a segmentation error in SYMPHONY occurring are coloured green. 
    All instances below the diagonal line benefit from warmstarting. }
    \label{fig:scatter_different_eps}
\end{figure}

We first analyse the categories in which the only difference between the source and target instances is the magnitude of~$\varepsilon$.
Using warmstarting in the category with source and target instance both resulting in UNSAT results, abbreviated as the UNSAT-UNSAT category, led to a substantial reduction in running time, of up to 97\%.
In addition, we were able to solve 170 instances that timed out in the baseline experiments.
Figure~\ref{fig:unsat_scatter} further illustrates this reduction in running time.
This category provides great potential in reducing the computational burden of certifying robustness against perturbations. 

For the baseline MILP instances, we observe relatively few SAT results; consequently, the number of warmstarted MILP instances with a SAT source instance is small (141), compared to 2021 instances with an UNSAT source instance.
Interestingly, most baseline MILP instances that are expected to be SAT result in timeouts.
Although SAT instances are generally considered easier to solve in state-of-the-art neural network verifiers, these verifiers often employ specialised techniques, such as adversarial attacks, to efficiently search for counterexamples.
SYMPHONY, as a general-purpose MILP solver, does not incorporate such techniques.
Therefore, SYMPHONY may not efficiently locate an adversarial example even when one exists, which may explain the timeouts observed for these instances.

The SAT--SAT category is the only category in which we observe an increase in running time when using warmstarting with \textsc{SYMPHONY} under the default warmstarting configuration of the tool;
this can also be observed from Figure \ref{fig:sat_scatter}.
This suggests that warmstarting from a smaller $\varepsilon$-value is not beneficial.
However, this observation should be interpreted with caution, as these SAT instances already require relatively short running times.
Indeed, in the SAT--TIMEOUT category, warmstarting enables solving target instances that previously timed out.

\subsection{Different \texorpdfstring{$x^*$}{input}.}
For the category in which the source and target instances differ only in the input image, while sharing the same network and perturbation radius~$\varepsilon$ 
(denoted as IMAGE), we observe a substantial potential reduction in running time when using warmstarting:
averaged over all networks we have considered, warmstarting leads to a reduction in running time of 85\%.
This improvement is also evident from Figure~\ref{fig:image_scatter}, which notably shows a large number of instances that timed out in the baseline experiments but are successfully solved when using warmstarting.

Notably, this reduction is achieved even though changing the input image alters the left-hand side of the MILP formulation, constituting a substantial 
modification of the optimisation problem.
In this work, we do not employ bound propagation techniques, which in state-of-the-art verifiers would further influence neuron bounds across different inputs;
this is an important consideration when transferring these insights to modern verification frameworks.
Nevertheless, our results indicate that warmstarting can be highly effective for this category.

\subsection{Different \texorpdfstring{$f_{\theta}$}{model}.}
Although we did not initially expect warmstarting across different neural networks to be beneficial, we nevertheless observed an average reduction in running time of 64\%.
As shown in Figure~\ref{fig:network_scatter}, this reduction is largely driven by instances that previously timed out in the baseline experiments but can be solved when using warmstarting.
For the remaining instances, the points are distributed relatively evenly around the diagonal, indicating that warmstarting benefits some MILP instances based on different networks, while increasing running time for others. 
Identifying which MILP features lead to performance improvements is an interesting direction for future work.

\section{Conclusions and Future Work}\label{sec:conclusions}

In this work, we have studied the effectiveness of solver-level warmstarting for neural network verification using MILP solver SYMPHONY.
Our results demonstrate that this form of warmstarting can lead to substantial efficiency gains, particularly for more general instance variations, in which properties differ beyond minor parameter changes.
Most notably, for instance pairs in which both the source and target instances are UNSAT (i.e., both are verified to be robust),
we observe reductions of up to 97\% in average running time, and, across all investigated instance pair categories, warmstarting enables the successful solving of 300 out of 474 instances that previously timed out. 
At the same time, our analysis highlights important limitations: warmstarting is not beneficial in all cases, and, in 301 out of 1128 individual cases, it even increased running time by on average 364\%.

These findings suggest several promising directions for future work.
In particular, it would be valuable to investigate how warmstarting, as explored here, can be integrated into state-of-the-art neural network verifiers (e.g., $\alpha, \beta$-CROWN \cite{wang2021beta}), which often rely on more advanced branching strategies, tighter relaxations and parallel solving techniques.
Using warmstarting in such verifiers may enable more informed initialisation of the algorithm for new instances, potentially also improving verification efficiency, as demonstrated in our study. 
In addition, future work could explore which features of MILP instances render warmstarting effective, thus enabling the automatic selection of instances for warmstarting.

\section*{Acknowledgements}
This research was
partially supported by the ELSA mobility grant, a project funded by the European Union, and by an Alexander-von-Humboldt Professorship in AI held by Holger Hoos.
Minghao Liu and Marta Kwiatkowska were supported by the EPSRC Prosperity Partnership FAIR (grant number EP/V056883/1)
and
ELSA: European Lighthouse on Secure and Safe AI project (grant agreement
No. 101070617 under UK guarantee).

\bibliographystyle{splncs04}
\bibliography{references}

@article{BosmanEtAl25,  
    author = {Bosman, Annelot W. and Berger, Aaron and Hoos, Holger H. and van Rijn, Jan N.},
    title = {Robustness Distributions in Neural Network Verification},
    year = {2025},
    journal = {Journal of Artificial Intelligence Research},
    pages={1--27},
    number ={20},
    volume ={83}
}

@article{ugare2023incremental,
  title={Incremental verification of neural networks},
  author={Ugare, Shubham and Banerjee, Debangshu and Misailovic, Sasa and Singh, Gagandeep},
  journal={Proceedings of the ACM on Programming Languages},
  volume={7},
  number={PLDI},
  pages={1920--1945},
  year={2023},
  publisher={ACM New York, NY, USA}
}

@article{banerjee2024input,
  title={Input-relational verification of deep neural networks},
  author={Banerjee, Debangshu and Xu, Changming and Singh, Gagandeep},
  journal={Proceedings of the ACM on Programming Languages},
  volume={8},
  number={PLDI},
  pages={1--27},
  year={2024},
  publisher={ACM New York, NY, USA}
}

@inproceedings{fischer2022shared,
  title={Shared certificates for neural network verification},
  author={Fischer, Marc and Sprecher, Christian and Dimitrov, Dimitar Iliev and Singh, Gagandeep and Vechev, Martin},
  booktitle={International Conference on Computer Aided Verification},
  pages={127--148},
  year={2022},
  organization={Springer}
}

@inproceedings{tjeng_evaluating_2019,
  author       = {Vincent Tjeng and
                  Kai Yuanqing Xiao and
                  Russ Tedrake},
  title        = {Evaluating Robustness of Neural Networks with Mixed Integer Programming},
  booktitle    = {Proceedings of the 7th International Conference on Learning Representations, {ICLR}},
  year         = {2019},
  pages = {1--11}

}

@article{de2021improved,
  title={Improved branch and bound for neural network verification via lagrangian decomposition},
  author={De Palma, Alessandro and Bunel, Rudy and Desmaison, Alban and Dvijotham, Krishnamurthy and Kohli, Pushmeet and Torr, Philip HS and Kumar, M Pawan},
  journal={arXiv preprint arXiv:2104.06718},
  year={2021}
}

@article{meng2022adversarial,
  title={Adversarial robustness of deep neural networks: A survey from a formal verification perspective},
  author={Meng, Mark Huasong and Bai, Guangdong and Teo, Sin Gee and Hou, Zhe and Xiao, Yan and Lin, Yun and Dong, Jin Song},
  journal={IEEE Transactions on Dependable and Secure Computing},
  year={2022},
  publisher={IEEE}
}

@article{naude2021artificial,
  title={Artificial Intelligence: Neither Utopian nor Apocalyptic Impacts Soon},
  author={Naud{\'e}, Wim},
  journal={Economics of Innovation and New Technology},
  volume={30},
  number={1},
  pages={1--23},
  year={2021},
  publisher={Taylor \& Francis}
}

@inproceedings{LeCunEtAl1998,
  author={Lecun, Y. and Bottou, L. and Bengio, Y. and Haffner, P.},
  booktitle={Proceedings of the IEEE}, 
  title={Gradient-based learning applied to document recognition}, 
  year={1998},
  pages={2278-2324},
//address = {unknown},
publisher ={IEEE}}

@article{zhu2024autonomous,
  title={Autonomous driving with spiking neural networks},
  author={Zhu, Rui-Jie and Wang, Ziqing and Gilpin, Leilani and Eshraghian, Jason},
  journal={Advances in Neural Information Processing Systems},
  year={2024}
}

@article{zhou2021reviewmedical,
  title={A review of deep learning in medical imaging: Imaging traits, technology trends, case studies with progress highlights, and future promises},
  author={Zhou, S Kevin and Greenspan, Hayit and Davatzikos, Christos and Duncan, James S and Van Ginneken, Bram and Madabhushi, Anant and Prince, Jerry L and Rueckert, Daniel and Summers, Ronald M},
  journal={Proceedings of the IEEE},
  volume={109},
  number={5},
  pages={820--838},
  year={2021},
  publisher={IEEE}
}

@article{szegedy2013intriguing,
  title={Intriguing properties of neural networks},
  author={Szegedy, Christian and Zaremba, Wojciech and Sutskever, Ilya and Bruna, Joan and Erhan, Dumitru and Goodfellow, Ian and Fergus, Rob},
  journal={arXiv preprint arXiv:1312.6199},
  year={2013}
}

@article{liu2016delving,
  title={Delving into transferable adversarial examples and black-box attacks},
  author={Liu, Yanpei and Chen, Xinyun and Liu, Chang and Song, Dawn},
  journal={arXiv preprint arXiv:1611.02770},
  year={2016}
}

@article{naseer2019cross,
  title={Cross-domain transferability of adversarial perturbations},
  author={Naseer, Muhammad Muzammal and Khan, Salman H and Khan, Muhammad Haris and Shahbaz Khan, Fahad and Porikli, Fatih},
  journal={Advances in Neural Information Processing Systems},
  volume={32},
  year={2019}
}

@inproceedings{waseda2023closer,
  title={Closer look at the transferability of adversarial examples: How they fool different models differently},
  author={Waseda, Futa and Nishikawa, Sosuke and Le, Trung-Nghia and Nguyen, Huy H and Echizen, Isao},
  booktitle={Proceedings of the IEEE/CVF Winter Conference on Applications of Computer Vision},
  pages={1360--1368},
  year={2023}
}

@inproceedings{ralphs2006duality,
  title={Duality and warm starting in integer programming},
  author={Ralphs, Ted and G{\"u}zelsoy, Menal},
  booktitle={The proceedings of the 2006 NSF design, service, and manufacturing grantees and research conference},
  volume={40},
  year={2006}
}

@inproceedings{wang2021beta,
  title={{Beta-CROWN}: Efficient bound propagation with per-neuron split constraints for complete and incomplete neural network verification},
  author={Wang, Shiqi and Zhang, Huan and Xu, Kaidi and Lin, Xue and Jana, Suman and Hsieh, Cho-Jui and Kolter, J Zico},
  booktitle={Advances in {{Neural Information Processing Systems (NeurIPS)}}},
  volume={34},
  year={2021}
}

@article{banerjee2024relational,
  title={Relational {DNN} verification with cross executional bound refinement},
  author={Banerjee, Debangshu and Singh, Gagandeep},
  journal={arXiv preprint arXiv:2405.10143},
  year={2024}
}

@inproceedings{katz2017reluplex,
  title={{Reluplex}: An Efficient {SMT} Solver for Verifying Deep Neural Networks},
  author={Katz, Guy and Barrett, Clark and Dill, David L and Julian, Kyle and Kochenderfer, Mykel J},
  booktitle={Proceedings of the 29th International Conference on Computer Aided Verification (CAV 2017)},
  year={2017},
  pages = {97--117}
}

@article{tzour2025mini,
  title={Mini-Batch Robustness Verification of Deep Neural Networks},
  author={Tzour-Shaday, Saar and Drachsler-Cohen, Dana},
  journal={Proceedings of the ACM on Programming Languages},
  volume={9},
  number={OOPSLA2},
  pages={2786--2814},
  year={2025},
  publisher={ACM New York, NY, USA}
}

@incollection{ralphs2005symphony,
  title={The SYMPHONY callable library for mixed integer programming},
  author={Ralphs, Ted K and G{\"u}zelsoy, Menal},
  booktitle={The next wave in computing, optimization, and decision technologies},
  pages={61--76},
  year={2005},
  publisher={Springer}
}

@article{dantzig1951maximization,
  title={Maximization of a linear function of variables subject to linear inequalities},
  author={Dantzig, George B},
  journal={Activity analysis of production and allocation},
  volume={13},
  pages={339--347},
  year={1951}
}

@inproceedings{liu2026exact,
  title={Exact verification of graph neural networks with incremental constraint solving},
  author={Liu, Minghao and Lu, Chia-Hsuan and Kwiatkowska, Marta},
  booktitle={International Symposium on Formal Methods},
  pages={641--662},
  year={2026},
  organization={Springer}
}

@article{elsaleh2026incremental,
  title={Incremental Neural Network Verification via Learned Conflicts},
  author={Elsaleh, Raya and Davis, Liam and Wu, Haoze and Katz, Guy},
  journal={arXiv preprint arXiv:2603.12232},
  year={2026}
}
\newpage
\appendix

\section{Tool Configuration Details}\label{app:tools}

In this work, we combine tools from both the neural network verification domain and the optimisation domain, as described in Section~\ref{sec:warmstarting_pipeline}. In the following, we describe the configuration of each tool used in our experimental pipeline.

First, we use VERONA\footnote{\url{https://github.com/ADA-research/VERONA}} as our experiment manager. We forked the repository and added a \emph{VerificationModule} that interfaces with both Marabou\footnote{\url{https://github.com/NeuralNetworkVerification/Marabou}} and SYMPHONY\footnote{\url{https://github.com/coin-or/SYMPHONY}}. 

To generate MILP encodings, we use Marabou through its command-line interface with the \texttt{----milp} flag. 
We additionally introduced a command-line option that exports the MPS formulation generated by Gurobi and terminates immediately afterwards. 
This modification allows us to obtain the MPS files required for our experiments, while avoiding changes that would cause unit tests to fail when building Marabou from source.

Finally, we use SYMPHONY to run both the baseline verification experiments and the warmstarting experiments. 
In both experiments, we use the default solver configuration.

\end{document}